\documentclass[11pt,a4paper]{article}
\usepackage[hyperref]{acl2018}
\usepackage{times}
\usepackage{latexsym}

\usepackage{url}

\aclfinalcopy 
\def\aclpaperid{1616} 

\usepackage{xspace}
\usepackage{mathtools}
\usepackage{amsmath}
\usepackage{amssymb}
\usepackage{color}
\usepackage{multirow}
\usepackage{tikz}
\usepackage{pifont}
\usepackage[inline]{enumitem}
\usetikzlibrary{positioning}
\usetikzlibrary{calc}
\usetikzlibrary{fit}
\usetikzlibrary{decorations.text}
\usetikzlibrary{shapes.misc}
\usetikzlibrary{shapes.geometric,arrows}
\usetikzlibrary{intersections}
\usepackage{colortbl}
\usepackage{mathrsfs}
\usepackage{bm}
\usepackage{booktabs}
\usepackage{pgfplots}
\usepackage{arydshln}
\usepackage{framed}

\newcommand{\eg}{e.g.\ }

\newcommand{\system}[2][]{\texttt{#2#1}\xspace}

\newcommand{\tabref}[2][]{Table#1~\ref{#2}\xspace}
\newcommand{\figref}[2][]{Figure#1~\ref{#2}\xspace}

\newcommand{\equref}[2][]{Equation#1~\ref{#2}}

\newcommand{\wtov}{\system{word2vec}}

\newcommand{\bb}[1]{\mathbb{#1}}
\newcommand{\R}{\bb{R}}
\newcommand{\mat}[2][]{\boldsymbol{#2}_{#1}}
\newcommand{\matfwd}[2][]{\overrightarrow{\bm{#2}}_{#1}}

\renewcommand{\vec}[2][]{\boldsymbol{#2}^{#1}}
\newcommand{\vecfwd}[2][]{\overrightarrow{\bm{#2}}^{#1}}
\newcommand{\vecrev}[2][]{\overleftarrow{\bm{#2}}^{#1}}
\newcommand{\vectilde}[2][]{\tilde{\bm{#2}}^{#1}}
\newcommand{\vectildefwd}[2][]{\overrightarrow{\tilde{\bm{#2}}}^{#1}}

\newcommand{\vecringfwd}[2][]{\overrightarrow{\mathring{\bm{#2}}}^{#1}}

\newcommand{\gru}{\textrm{GRU}}
\newcommand{\grufwd}{\overrightarrow{\textrm{GRU}}}

\newcommand{\softmax}{\textrm{softmax}}
\newcommand{\xentropy}{\ensuremath{\operatorname{XEntropy}}\xspace}

\newcommand{\T}{\mathstrut\scriptscriptstyle\top}

\newcommand{\ent}[1][]{\system[#1]{EntNet}}

\newcommand{\lstm}[1][]{\system[#1]{LSTM}}

\newcommand{\bigru}[1][]{\system[#1]{Bi-GRU}}

\newcommand{\dssm}{\system{DSSM}}
\newcommand{\msap}{\system{MSAP}}
\newcommand{\hcm}{\system{HCM}}
\newcommand{\tbmihaylov}{\system{TBMIHAYLOV}}

\newcommand{\RNum}[1]{\uppercase\expandafter{\romannumeral #1\relax}}

\title{Narrative Modeling with Memory Chains and Semantic Supervision}

\author{Fei Liu \qquad Trevor Cohn \qquad Timothy Baldwin \\
         School of Computing and Information Systems \\ The University of Melbourne \\ Victoria, Australia\\
         {\tt {fliu3@student.unimelb.edu.au}} \\
         {\tt {t.cohn@unimelb.edu.au}\,\,\, {tb@ldwin.net}}}

\date{}

\begin{document}
\maketitle
\begin{abstract}
Story comprehension requires a deep semantic understanding of the
narrative, making it a challenging task.
Inspired by previous studies on ROC Story Cloze Test, we propose a novel
method, tracking various semantic aspects with external neural memory
chains while encouraging each to focus on a particular semantic aspect.
Evaluated on the task of story ending prediction, our model demonstrates superior performance to a collection of competitive baselines, setting a new state of the art. \footnote{Code available at \url{http://github.com/liufly/narrative-modeling}.}
\end{abstract}

\section{Introduction}

Story narrative comprehension has been a long-standing challenge in
artificial intelligence
\cite{Winograd:1972,Turner:1994,Schubert+:2000}. The difficulties of
this task arise from the necessity for understanding not only
narratives, but also commonsense and normative social behaviour
\cite{Charniak:1972}. Of particular interest in this paper is the work
by \newcite{Mostafazadeh+:2016} on understanding commonsense stories in
the form of a Story Cloze Test: given a short story, we must predict the
most coherent sentential ending from two options (e.g.\ see \figref{fig:example}).

Many attempts have been made to solve this problem, based either on linear classifiers with handcrafted features \cite{Schwartz+:2017,Chaturvedi+:2017}, or representation learning via deep learning models \cite{Mihaylov+:2017,Bugert+:2017,Mostafazadeh+:2017}. 
Another widely used component of competitive systems is language model-based features, which require training on large corpora in the story domain.

The current state-of-the-art approach of \newcite{Chaturvedi+:2017} is based on understanding the context from three perspectives: (1) event sequence, (2) sentiment trajectory, and (3) topic consistency. \newcite{Chaturvedi+:2017} adopt external tools to recognise relevant aspect-triggering words, and manually design features to incorporate them into the classifier. 
While identifying triggers has been made easy by the use of various linguistic resources, crafting such features is time consuming and requires domain-specific knowledge along with repeated experimentation.

\begin{figure}
\small
\begin{framed}
\textbf{Context}: Sam loved his old belt. He matched it with everything. Unfortunately he gained too much weight. It became too small.\\
\textbf{Coherent Ending}: Sam went on a diet.\\
\textbf{Incoherent Ending}: Sam was happy.
\end{framed}
\caption{Story Cloze Test example.}
\label{fig:example}
\end{figure}

Inspired by the argument for tracking the dynamics of events, sentiment and topic, we propose to address the issues identified above with multiple external memory chains, each responsible for a single aspect. Building on Recurrent Entity Networks (\ent[s]), a superior framework to \lstm[s] demonstrated by \newcite{Mikael+:2017} for reasoning-focused question answering and cloze-style reading comprehension, we introduce a novel multi-task learning objective, encouraging each chain to focus on a particular aspect. While still making use of external linguistic resources, we do not extract or design features from them but rather utilise such tools to generate labels. The generated labels are then used to guide training so that each chain focuses on tracking a particular aspect. At test time, our model is free of feature engineering such that, once trained, it can be easily deployed to unseen data without preprocessing. Moreover, our approach also differs in the lack of a ROC Stories language model component, eliminating the need for large, domain-specific training corpora.

Evaluated on the task of Story Cloze Test, our model outperforms a collection of competitive baselines, achieving state-of-the-art performance under more modest data requirements.


\section{Methodology}

In the story cloze test, given a story of length $L$, consisting of a sequence of context sentences $\mathbf{c} = \{c_1, c_2, \ldots, c_L\}$, we are interested in predicting the coherent ending to the story out of two ending options $o_1$ and $o_2$. Following previous studies \cite{Schwartz+:2017}, we frame this problem as a binary classification task. Assuming $o_1$ and $o_2$ are the logical and inconsistent endings respectively with their associated labels being $y_1$ and $y_2$, we assign $y_1 = 1$ and $y_2 = 0$. At test time, given a pair of possible endings, the system returns the one with the higher score as the prediction. In this section, we first describe the model architecture and then detail how we identify aspect-triggering words in text and incorporate them in training.


\subsection{Proposed Model}

First, to take neighbouring contexts into account, we process context sentences and ending options at the word level with a bi-directional gated recurrent unit (``\bigru'': \newcite{Chung+:2014}): $\vecfwd{h}_{i} = \grufwd(\vec{w}_{i}, \vecfwd{h}_{i-1})$ where $\vec{w}_{i}$ is the embedding of the $i$-th word in the story or ending option. The backward hidden representation $\vecrev{h}_{i}$ is computed analogously but with a different set of $\gru$ parameters. We simply take the sum $\vec{h}_{i} = \vecfwd{h}_i + \vecrev{h}_i$ as the representation at time $i$. An ending option, denoted $\vec{o}$, is represented by taking the sum of the final states of the same \bigru over the ending option word sequence. 

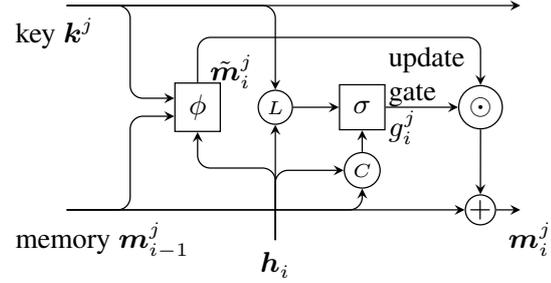
\begin{figure}[t]
\begin{center}
\resizebox{\columnwidth}{!}{
\begin{tikzpicture}

\node[align=center,minimum size=0.5cm] at (0.0, 0.0) (wt) {\footnotesize  $\vec{h}_{i}$};
\node[align=center] [above left = 0.2 and 2 of wt] (ht_1) {};
\node[align=center] [above = 2.0 of ht_1] (kj) {};

\node[draw,align=center,minimum size=0.5cm] [above left = 1.2 and 0.3 of wt] (phi) {\footnotesize $\phi$};



\draw [rounded corners,->,>=stealth,name path=k_path] (kj.east) -- node [at start,below,shift={(-0.7,0)},anchor=north west] (kj_label) {\footnotesize key $\vec[j]{k}$} ($(kj.east) + (5,0.0)$);


\draw [rounded corners,->,>=stealth,name path=w_path] (wt.north) -- ($(wt.north) + (0,0.8)$) -| (phi.south);

\draw [rounded corners,->,>=stealth] (ht_1.east) -- ($(ht_1.east) + (0.7,0.0)$) |- ($(phi.west) + (0.0, -0.1)$);

\draw [rounded corners,->,>=stealth] (kj.east) -- ($(kj.east) + (0.7,0.0)$) |- ($(phi.west) + (0.0, 0.1)$);



\node[draw,align=center,minimum size=0.5cm] [right = 1.3 of phi] (sigma) {\footnotesize $\sigma$};

\node[draw,circle,align=center,inner sep=1.5pt] [above = 1.28 of wt] (kjwt) {\tiny $L$};

\draw [rounded corners,->,>=stealth] (wt.north) -- (kjwt.south);

\draw [rounded corners,->,>=stealth] (kj.east) -| (kjwt.north);

\draw [rounded corners,->,>=stealth] (kjwt.east) -- (sigma.west);

\node[draw,circle,align=center,inner sep=1.5pt] [below = 0.24 of sigma] (ht_1wt) {\tiny $C$};

\draw [rounded corners,->,>=stealth] (wt.north) |- (ht_1wt.west);

\draw [rounded corners,->,>=stealth] (ht_1.east) -| (ht_1wt.south);

\draw [rounded corners,->,>=stealth] (ht_1wt.north) -- (sigma.south);


\node[draw,circle,align=center,inner sep=1.5pt] [right = 0.8 of sigma] (g) {\footnotesize $\odot$};

\draw [rounded corners,->,>=stealth] (sigma.east) -- node [at start,below,shift={(-0.1,0.8)},anchor=north west,text width=1] (g_label1) {\footnotesize update} node [at start,below,shift={(-0.1,0.4)},anchor=north west,text width=1] (g_label1) {\footnotesize gate} node [at start,below,shift={(-0.1,0.1)},anchor=north west,text width=1] (g_label1) {\footnotesize $g_i^{j}$} (g.west);

\draw [rounded corners,->,>=stealth] (phi.north) --  node [at start,above,shift={(0.0,-0.2)},anchor=south west] (memory_candidate_label) {\footnotesize $\vectilde[j]{m}_{i}$} ($(phi.north) + (0,0.5)$) -| (g.north);


\node[draw,circle,align=center,inner sep=0pt] [right = 4.387 of ht_1] (ht) {\footnotesize $+$};

\draw [rounded corners,->,>=stealth,name path=h_path] (ht_1.east) -- node [at start,below,shift={(-0.7,0)},anchor=north west] (ht_1_label) {\footnotesize memory $\vec[j]{m}_{i-1}$} (ht.west);

\draw [rounded corners,->,>=stealth] (g.south) -- (ht.north);

\draw [rounded corners,->,>=stealth] (ht.east) -- node [at start,below,shift={(0.0,0.0)},anchor=north west] (ht_label) {\footnotesize $\vec[j]{m}_i$} ($(ht_1.east) + (5,0)$);

\end{tikzpicture}
}
\end{center}
\caption{Illustration of \ent with a single memory chain at time $i$. $\phi$ and $\sigma$ represent \equref[s]{equ:phi} and \ref{equ:gate}, while circled nodes $L$, $C$, $\odot$ and $+$ depict the location, content terms, Hadamard product, and addition, resp.}
\label{fig:model}
\end{figure}

This representation is then taken as input to a Recurrent Entity Network
(``\ent'': \newcite{Mikael+:2017}), capable of tracking the state of the
world with external memory. Developed in the context of machine reading
comprehension, \ent maintains a number of ``memory chains'' --- one for
each entity --- and dynamically updates the representations of them as
it progresses through a story. Here, we borrow the concept of ``entity''
and use each chain to track a unique aspect. An illustration of \ent is
provided in \figref{fig:model}.

\paragraph{Memory update candidate.} At every time step $i$, the internal memory update candidate $\vectildefwd[j]{m}_{i}$ for the $j$-th memory chain is computed as:
\begin{equation}
\vectildefwd[j]{m}_{i} = \phi(\matfwd{U}\vecfwd[j]{m}_{i-1} + \matfwd{V}\vec[j]{k} + \matfwd{W}\vec{h}_{i})
\label{equ:phi}
\end{equation}
where $\vec[j]{k}$ is the embedding for the $j$-th entity (key), $\matfwd{U}$, $\matfwd{V}$ and $\matfwd{W}$ are trainable parameters, and $\phi$ is the parametric ReLU \cite{He+:2015}.

\paragraph{Memory update.} The update of each memory chain is controlled by a gating mechanism:
\begin{align}
\overrightarrow{g}_{i}^{j} &= \sigma(\vec{h}_{i}\cdot\vec[j]{k} + \vec{h}_{i}\cdot\vecfwd[j]{m}_{i-1}) \label{equ:gate}\\
\vecringfwd[j]{m}_{i} &= \vecfwd[j]{m}_{i-1} + \overrightarrow{g}_{i}^{j}\odot\vectildefwd[j]{m}_{i}
\end{align}
where $\odot$ denotes Hadamard product (resulting in the unnormalised memory representation $\vecringfwd[j]{m}_{i}$), and $\sigma$ is the sigmoid function. The gate $\overrightarrow{g}_{i}^{j}$ controls how much the memory chain should be updated, a decision factoring in two elements: (1) the ``location'' term, quantifying how related the current input $\vec{h}_i$ (the output of the \bigru at time $i$) is to the key $\vec[j]{k}$ of the entity being tracked; and (2) the ``content'' term, measuring the similarity between the input and the current state $\vecfwd[j]{m}_{i-1}$ of the entity tracked by the $j$-th memory chain.

\paragraph{Normalisation.} We normalise each memory representation: $\vecfwd[j]{m}_i = \vecringfwd[j]{m}_{i}/\|\vecringfwd[j]{m}_{i}\|$ where $\|\vecringfwd[j]{m}_{i}\|$ denotes the Euclidean norm of $\vecringfwd[j]{m}_{i}$. In doing so, we allow the model to forget in the sense that, as $\vecfwd[j]{m}_i$ is constrained to be lying on the surface of the unit sphere, adding new information carried by $\vectildefwd[j]{m}_i$ and then performing normalisation inevitably causes forgetting in that the cosine distance between the original and updated memory decreases.

\paragraph{Bi-directionality.} In contrast to \ent, we apply the above steps in both directions, scanning a story both forward and backward, with arrows denoting the processing direction. $\vecrev[j]{m}_i$ is computed analogously to $\vecfwd[j]{m}_i$ but with a different set of parameters, and 
$\vec[j]{m}_i = \vecfwd[j]{m}_i + \vecrev[j]{m}_i$.
We further define $g_{i}^{j}$ to be the average of the gate values of both directions at time $i$ for the $j$-th chain:
\begin{equation}
g_{i}^{j} = (\overrightarrow{g}_{i}^{j} + \overleftarrow{g}_{i}^{j})/2
\label{equ:semanticpred}
\end{equation}
which will be used for semantic supervision.

\paragraph{Final classifier.} The final prediction $\hat{y}$ to an ending option given its context story is performed by incorporating the last states of all memory chains in the form of a weighted sum $\vec{u}$:
\begin{align}
p^{j} &= \softmax((\vec[j]{k})^{\T}\mat[att]{W}\vec{o})\\
\vec{u} &= \sum_{j} p^{j}\vec[j]{m}_T
\end{align}
where $T$ denotes the total number of words in a story and $\mat[att]{W}$ is a trainable weight matrix. 
We then transform $\vec{u}$ to get the final prediction:
\begin{equation}
\hat{y} = \sigma(\mat{R}\phi(\mat{H}\vec{u} + \vec{o}))
\label{equ:prediction}
\end{equation}
where $\mat{R}$ and $\mat{H}$ are trainable weight matrices.

\subsection{Semantically Motivated Memory Chains}
\label{sec:semanticsupervision}

In order to encourage each chain to focus on a particular aspect, we supervise the activation of each update gate (\equref{equ:gate}) with binary labels. 
Intuitively, for the sentiment chain, a sentiment-bearing word (\eg \textit{amazing}) receives a label of $1$, allowing the model to open the gate and therefore change the internal state relevant to this aspect, while a neutral one (\eg \textit{school}) should not trigger the activation with an assigned label of $0$.
To achieve this, we use three memory chains to independently track each of: (1) event sequence, (2) sentiment trajectory, and (3) topical consistency. To supervise the memory update gates of these chains, we design three sequences of binary labels: $\mathbf{l}^{j} = \{l_1^{j}, l_2^{j}, \ldots, l_T^{j}\}$ for $j \in [1,3]$ representing event, sentiment, and topic, and $l_i^j \in \{0, 1\}$. The label at time $i$ for the $j$-th aspect is only assigned a value of $1$ if the word is a trigger for that particular aspect: $l_i^{j} = 1$; otherwise $l_i^{j} = 0$. During training, we utilise such sequences of binary labels to supervise the memory update gate activations of each chain. Specifically, each chain is encouraged to open its gate only when it sees a trigger bearing information semantically sensitive to that particular aspect. Note that at test time, we do not apply such supervision. This effectively becomes a binary tagging task for each memory chain independently and results in individual memory chains being sensitive to only a specific set of triggers bearing one of the three types of signals.

While largely inspired by \newcite{Chaturvedi+:2017}, our approach differs in how we extract and use such features.
Also note that, while still making use of external linguistic resources to detect trigger words, our approach lets the memory chains decide how such words should influence the final prediction, as opposed to the handcrafted conditional probability features of \newcite{Chaturvedi+:2017}.

\begin{table}
\resizebox{\columnwidth}{!}{
\begin{tabular}{l|c@{\hskip 2mm}c@{\hskip 2mm}c@{\hskip 2mm}c@{\hskip 2mm}c@{\hskip 2mm}c@{\hskip 2mm}c}
                         & Ricky & fell & while & hiking & in & the & woods \\
\midrule
$\mathbf{l}^{Event}$     &   0   &   1  &   0   &    1   & 0  &  0  & 1     \\
$\mathbf{l}^{Sentiment}$ &   0   &   1  &   0   &    0   & 0  &  0  & 0     \\
$\mathbf{l}^{Topic}$     &   1   &   1  &   0   &    1   & 0  &  0  & 1     \\
\end{tabular}
}
\caption{An example of the linguistically motivated memory chain supervision binary labels.}
\label{tbl:chainlabel}
\end{table}

To identify the trigger words, we use external linguistic tools, and mark trigger words for each aspect with a label of $1$. An example is presented in \tabref{tbl:chainlabel}, noting that the same word can act as trigger for multiple aspects.

\paragraph{Event sequence.} We parse each sentence into its FrameNet representation with \system{SEMAFOR} \cite{Das+:2010}, and identify each frame target (word or phrase tokens evoking a frame).

\paragraph{Sentiment trajectory.} Following \newcite{Chaturvedi+:2017}, we utilise a pre-compiled list of sentiment words \cite{Liu+:2005}. To take negation into account, we parse each sentence with the \system{Stanford Core NLP} dependency parser \cite{Manning+:2014,Chen+:2014} and include negation words as trigger words.

\paragraph{Topical consistency.} We process each sentence with the \system{Stanford Core NLP} POS tagger and identify nouns and verbs, following \newcite{Chaturvedi+:2017}.

\subsection{Training Loss}

In addition to the cross entropy loss of the final prediction of right/wrong endings, we also take into account the memory update gate supervision of each chain by adding the second term.
More formally, the model is trained to minimise the loss: 
\begin{equation*}
\label{equ:loss}
\mathcal{L} = \xentropy(y,\hat{y}) + \alpha\sum_{i,j}\xentropy(l_i^j,g_i^{j})
\end{equation*}
where $\hat{y}$ and $g_i^{j}$ are defined in \equref[s]{equ:prediction} and \ref{equ:semanticpred} respectively, $y$ is the gold label for the current ending option $o$, and $l_i^j$ is the semantic supervision binary label at time $i$ for the $j$-th memory chain. In our experiments, we empirically set $\alpha$ to 0.5 based on development data.


\section{Experiments}

\subsection{Experimental Setup}

\paragraph{Dataset.}
To test the effectiveness of our model, we employ the Story Cloze Test dataset of \newcite{Mostafazadeh+:2016}. The development and test set each consist of 1,871 4-sentence stories, each with a pair of ending options.
Consistent with previous studies, we split the development set into a training and validation set (for early stopping),
resulting in 1,683 and 188 in each set, resp. Note that while
most current approaches make use of the much larger training set,
comprised of 100K 5-sentence ROC stories (with coherent endings only,
also released as part of the dataset) to train a language model, we make no use of this data.

\paragraph{Model configuration.}
We initialise our model with \wtov embeddings (300-D, pre-trained on 100B Google News articles, not updated during training: \newcite{Mikolov+:2013a,Mikolov+:2013b}). In addition to the three supervised chains, we also add a ``free'' chain, unconstrained to any semantic aspect. 

Training is carried out over $200$ epochs with the FTRL optimiser \cite{Mcmahan+:2013} and a batch size of $128$ and learning rate of $0.1$. We use the following hyper-parameters for weight matrices in both directions: $\mat{R} \in \R^{300\times1}$, $\mat{H}$, $\mat{U}$, $\mat{V}$, $\mat{W}$ are all matrices of size $\R^{300\times300}$, and hidden size of the \bigru is $300$. Dropout is applied to the output of $\phi$ in the final classifier (\equref{equ:prediction}) with a rate of $0.2$. Moreover, we employ the technique introduced by \newcite{Gal+:2016} where the same dropout mask is applied at every step to the input $\vec{w}_{i}$ to the \bigru and the input $\vec{h}_{i}$ to the memory chains with rates of $0.5$ and $0.2$ respectively. Lastly, to curb overfitting, we regularise the last layer (\equref{equ:prediction}) with an $L_2$ penalty on its weights: $\lambda\|\mat{R}\|$ where $\lambda = 0.001$.

\paragraph{Evaluation.}
Following previous studies, we evaluate the performance in terms of coherent ending prediction accuracy: $\frac{\#\text{correct}}{\#\text{total}}$. Here, we report the average accuracy over 5 runs with different random seeds. Echoing \newcite{Melis+:2017}, we also observe some variance in model performance even if training is carried out with the same random seed, which is largely due to the non-deterministic ordering of floating-point operations in our environment (\system{Tensorflow} \cite{tensorflow} with a single GPU). Therefore, to account for the randomness, we further train our model 5 times for each random seed and select the model with the best validation performance.

We benchmark against a collection of strong baselines, including the top-3 systems of the 2017 LSDSem workshop shared task \cite{Mostafazadeh+:2017}: \msap \cite{Schwartz+:2017}, \hcm \cite{Chaturvedi+:2017}\footnote{We take the performance reported in a subsequent paper, $3.2\%$ better than the original submission to the shared task.}, and \tbmihaylov \cite{Mihaylov+:2017}. The first two primarily rely on a simple logistic regression classifier, both taking linguistic and probability features from a ROC Stories domain-specific neural language model. \tbmihaylov is \lstm-based; we also include \dssm \cite{Mostafazadeh+:2016}. We additionally implement a bi-directional \ent \cite{Mikael+:2017} with the same hyper-parameter settings as our model and no semantic supervision.\footnote{\ent is also subject to the same model selection criterion described above.}

\subsection{Results and Discussion}

The experimental results are shown in \tabref{tbl:performance}. 

\begin{table}
\center
\begin{tabular}{lcc}
\toprule
Model & Acc. & SD \\
\midrule
\dssm               & 58.5 & --- \\
\tbmihaylov         & 72.8 & --- \\
\msap               & 75.2 & --- \\
\hcm                & 77.6 & --- \\
\ent$\dagger$       & 77.6 & $\pm 0.5$ \\
Our model$\dagger$ & \textbf{78.5} & $\pm 0.5$ \\
\bottomrule
\end{tabular}
\caption{Performance on the Story Cloze test set. \textbf{Bold} = best
  performance, $\dagger$ =  ave.\ accuracy over 5 runs, SD = standard
  deviation, ``---'' = not reported.}
\label{tbl:performance}
\end{table}

\paragraph{State-of-the-art results.} Our model outperforms a collection of strong baselines, setting a new state of the art. Note that this is achieved without any external linguistic resources at test time or
domain-specific language model-based probability features, highlighting the effectiveness of the proposed model.

\paragraph{\ent vs.\ our model.} We see clear improvements of the proposed method over \ent, an absolute gain of $0.9\%$ in accuracy. This validates the hypothesis that encouraging each memory chain to focus on a unique, well-defined aspect is beneficial.

\begin{figure}
	\centering
	\resizebox{\columnwidth}{!}{%
	\input{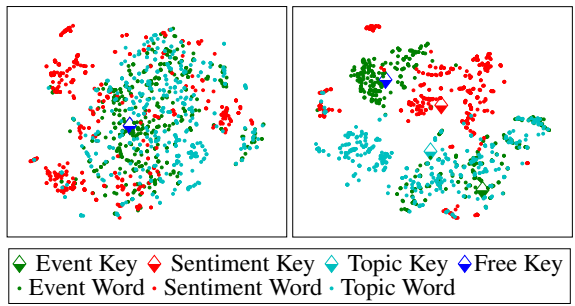}%
	}
	\caption{t-SNE scatter plot of aspect-triggering words (output representations $\vec{h}_i$ by \bigru, 500 randomly sampled from each aspect) and the learned keys. \ent (left) vs.\ our model (right). Best viewed in colour.}
	\label{fig:tsne}
\end{figure}

\paragraph{Discussion.} To better understand the benefits of the proposed method, we visualise the learned word representations (output representation of the \bigru, $\vec{h}_i$) and keys between \ent and our model in \figref{fig:tsne}. With \ent, while sentiment words form a loose cluster, words bearing event and topic signal are placed in close proximity and are largely indistinguishable.
With our model, on the other hand, the degree of separation is much clearer. The intersection of a small portion of the event and topic groups is largely due to the fact that both aspects include verbs. Another interesting perspective is the location of the automatically learned keys (displayed as diamonds): while all the keys with \ent end up overlapping, indicating little difference among them, the keys learned by our method demonstrate semantic diversity, with each placed within its respective cluster. Note that the free key is learned to complement the event key, a difficult challenge given the two disjoint clusters.

\paragraph{Ablation study.} We further perform a ablation study to analyse the utility of each component in \tabref{tbl:ablation}. The uni-directional variant, with its performance down to $77.8$, is inferior to its bi-directional cousin. Removing all semantic supervision (resulting in 4 free memory chains) is also damaging to the accuracy (dropping to $77.6$). Among the different types of semantic supervision, we see various degrees of performance degradation, with removing sentiment trajectory being the most detrimental, reflecting its value to the task. Interestingly, we observe improvement when removing event sequence supervision from consideration. We suspect that this is mainly due to the noise introduced by the rather inaccurate FrameNet representation output of \system{SEMAFOR} ($F_1 = 61.4\%$ in frame identification as reported in \newcite{Das+:2010}). While it is possible that replacing \system{SEMAFOR} with the SemLM approach (with the extended frame definition to include explicit discourse markers, \eg \textit{but}) in \newcite{Peng+:2016} and \newcite{Chaturvedi+:2017} may potentially reduce the amount of noise, we leave this exercise for future work.

\begin{table}
\center
\begin{tabular}{lcc}
\toprule
Model & Acc. & SD \\
\midrule
Our model & 78.5 & $\pm 0.5$ \\
---bi-directionality & 77.8 & $\pm 0.7$ \\
---all semantic supervision & 77.6 & $\pm 0.5$ \\
\hskip 1em ---event sequence & \textbf{78.7} & $\pm 0.2$ \\
\hskip 1em ---sentiment trajectory & 75.9 & $\pm 0.4$ \\
\hskip 1em ---topical consistency & 77.3 & $\pm 0.4$ \\
\hskip 1em ---free chain & 77.0 & $\pm 0.4$ \\
\bottomrule
\end{tabular}
\caption{Ablation study. Average accuracy over 5 runs on the Story Cloze test set (all subject to the same model selection criterion as our model). \textbf{Bold}: best performance. SD: standard deviation.}
\label{tbl:ablation}
\end{table}


\section{Conclusion}

In this paper, we have proposed a novel model for tracking various semantic aspects with external memory chains.
While requiring less domain-specific training data, our model achieves
state-of-the-art performance on the task of ROC Story Cloze ending
prediction, beating a collection of strong baselines.


\section*{Acknowledgments}

We thank the anonymous reviewers for their valuable feedback, and gratefully acknowledge the support of Australian Government Research Training Program Scholarship. This work was also supported in part by the Australian Research Council.

\bibliography{acl2018}
\bibliographystyle{acl_natbib}

\end{document}